\documentclass[letterpaper, 10 pt, conference]{ieeeconf}  

\usepackage{cite}
\usepackage{amsmath}
\usepackage{graphicx}
\usepackage{float} 
\usepackage{subfigure}
\usepackage{overpic}
\usepackage{wrapfig}
\usepackage{footmisc}

\IEEEoverridecommandlockouts                              

\title{\LARGE \bf
Efficient Implicit Neural Reconstruction Using LiDAR
}

\author{Dongyu Yan$^{1}$, Xiaoyang Lyu$^{2}$, Jieqi Shi$^{3}$, and Yi Lin$^{4}$
\thanks{$^{1}$Dongyu Yan is with School of Mechanical Engineering and Automation, Harbin Institute of Technology (Shenzhen).
        {\tt\small 21s053072@stu.hit.edu.cn}}
\thanks{$^{2}$Xiaoyang, Lyu is with Department of Electrical and Electronic Engineering, The University of Hong Kong.
        {\tt\small xylyu@eee.hku.hk}}
\thanks{$^{3}$Jieqi Shi is with the Department of Electronic and Computer Engineering, Hong Kong University of Science and Technology.
        {\tt\small {jshias}@connect.ust.hk}}
\thanks{$^{4}$Yi Lin is with Dji Co.
        {\tt\small {ylinax}@connect.ust.hk}}
}

\begin{document}

\maketitle
\thispagestyle{empty}
\pagestyle{empty}

\begin{abstract}

  Modeling scene geometry using implicit neural representation has revealed its advantages in accuracy, flexibility, and low memory usage.
  Previous approaches have demonstrated impressive results using color or depth images but still have difficulty handling poor light conditions and large-scale scenes.
  Methods taking global point cloud as input require accurate registration and ground truth coordinate labels, which limits their application scenarios.
  In this paper, we propose a new method that uses sparse LiDAR point clouds and rough odometry to reconstruct fine-grained implicit occupancy field efficiently within a few minutes.
  We introduce a new loss function that supervises directly in 3D space without 2D rendering, avoiding information loss.
  We also manage to refine poses of input frames in an end-to-end manner, creating consistent geometry without global point cloud registration.
  As far as we know, our method is the first to reconstruct implicit scene representation from LiDAR-only input.
  Experiments on synthetic and real-world datasets, including indoor and outdoor scenes, prove that our method is effective, efficient, and accurate, obtaining comparable results with existing methods using dense input.

\end{abstract}

\section{INTRODUCTION}

   Research on 3D reconstruction and scene representation using implicit neural representation has received extensive attention.
  Representing volume density field\cite{sucar2021imap}, signed distance function (SDF)\cite{park2019deepsdf, wang2021neus, ortiz2022isdf} or occupancy field\cite{mescheder2019occupancy, oechsle2021unisurf, zhu2021nice} with a neural network, researchers manage to reconstruct high-quality 3D models with high resolution and low memory cost.
  Different input data have been used to optimize the implicit representation, including RGB images, depth images, and point clouds.
  Following NeRF\cite{mildenhall2020nerf}, image-based implicit neural reconstruction methods\cite{wang2021neus, oechsle2021unisurf} use volume rendering to project 3D scenes to 2D and supervise with 2D photometric loss.
  However, RGB-only input can cause ambiguity due to occlusion, resulting in limited precision and noisy geometry.
  To overcome the limitations of RGB-only methods, some works\cite{sucar2021imap, zhu2021nice, ortiz2022isdf} utilize depth information from multi-view stereo or depth cameras to assist supervision and avoid ambiguity.
  Nevertheless, such methods still add supervision to 2D depth images by volume rendering, resulting in weak supervision.
  Meanwhile, the use of color and depth cameras also require suitable light condition and has difficulty in large-scale and outdoor scenes.
  There are also methods that reconstruct directly using 3D global point cloud, demonstrating impressive results in accuracy\cite{mescheder2019occupancy, chen2019learning, peng2020convolutional}.
  However, these methods face other significant challenges before they can be widely applied.
  First, these reconstruction methods usually require ground truth spatial attributes as supervision.
  Also, the raw point cloud frames must be registered globally into world coordinate with accurate poses to avoid inconsistency, which introduces errors and complexity to the system.
  In the end, classic methods usually take hours to build an implicit representation, which lacks timeliness.
  The above problems prevent these methods from being widely used in various scenarios.
  
  \begin{figure}[t]
  \centering
  \includegraphics[width=0.48\textwidth]{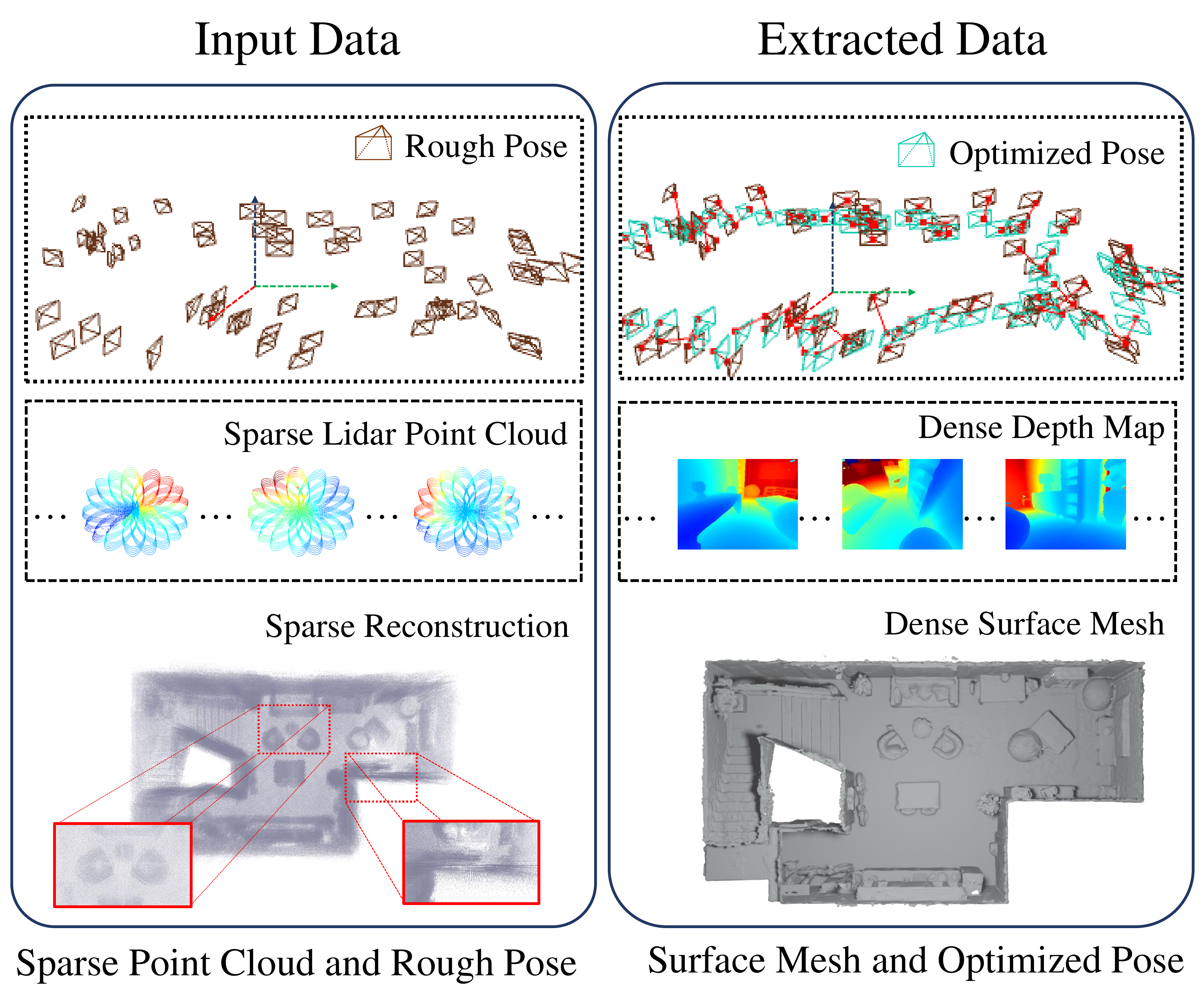}
  \vspace{-6mm}
  \caption{
 Our method takes sparse LiDAR point clouds as input and outputs a dense occupancy field using implicit representation.
    We propose a new direct supervision method to handle the sparsity of the input data and refine the initial rough poses in a joint optimization module.
  }
  \label{pipeline}
  \vspace{-6mm}
\end{figure}

 In this work, we explore the problem of optimizing an implicit occupancy field with only sparse LiDAR point clouds and set up a system that ensures accuracy, efficiency, robustness, and convenience at the same time.
  We point out the difficulty of the problem is that one LiDAR frame only contains 
  around $5\%$ depth data of a depth image.
  Such a difference indicates that depth rendering may not be able to provide enough geometry information for training and inspires us to seek a new way of supervision that better adapts to the sparsity of LiDAR point cloud.
  Leveraging the property of laser traveling, we propose to supervise directly in 3D space and manage to get rid of volume rendering and ground truth labels.
  With our object-thickness assumption, the new direct supervision achieves unbiased and occlusion-aware.
  Besides, we manage to free our method from cumbersome global registration by treating each frame individually, which avoids blending them into a global point cloud.
  Such modification enables us to use rough poses provided by odometry as initial and refine them along with the optimization of the scene representation\cite{wang2021nerf, lin2021barf}.
  Moreover, we refer to Instant Neural Graphics Primitives (Instant-NGP)\cite{muller2022instant} and use a multi-resolution hash encoder to accelerate the reconstruction process and provide fine-grained local and global geometry efficiently within a few minutes. 

  In summary, we present a method that takes sparse LiDAR point clouds and rough odometry as input and optimize an implicit occupancy field for reconstruction.
  Our method works efficiently and achieves good reconstruction results much faster than rendering-based methods.
  Also, due to the adaptability of LiDAR measurement, our method is suitable for a wide range of scenarios, including large-scale outdoor scenes.
  Our main contributions are as follows:
\begin{itemize}
  \item[$\bullet$] We propose a new loss function that enables dense implicit reconstruction from sparse LiDAR inputs;
  \item[$\bullet$] We add pose refinement along with scene optimization to further decrease initial pose error and create consistent geometry;
  \item[$\bullet$] Our implemented reconstruction method is efficient and only consumes a few minutes per scene.
  
\end{itemize}

\section{RELATED WORK}

\paragraph{Classical Method}
  The field of 3D reconstruction has been well explored using classical methods.
  Researchers reconstruct 3D models from different kinds of input data, including RGB-only\cite{schonberger2016structure, engel2017direct}, RGB-D\cite{izadi2011kinectfusion, whelan2013robust, dai2017bundlefusion} and LiDAR point cloud\cite{oleynikova2017voxblox}.
  RGB-only methods exploit photometric consistency between images and optimize 3D information under geometry constraints.
  Methods utilizing depth data from depth camera or LiDAR directly obtain 3D information and build a global point cloud map.
  However, noisy depth measurement may create inconsistent geometry.
  To overcome this, fusion-based methods\cite{izadi2011kinectfusion, dai2017bundlefusion} propose to maintain a global volumetric map containing geometry information such as signed distance or occupancy, which can be used to fuse and refine noisy depth data.
  By this means, 3D reconstruction tasks can be processed globally by optimization in a continuous way.
  
\paragraph{Learning-based Method}
Classical methods may lose efficacy when faced with poor light conditions or intense depth noise.
  Learning-based methods utilize photometric and geometric priors to perform robust reconstruction to reduce these effects.
  Some works\cite{detone2018superpoint, sarlin2020superglue, sarlin2021back} utilize learned features to replace hand-craft features for better correlation detection.
  Others abandon the traditional feature matching process and build a cost volume to mimic the multi-view stereo matching using 3D CNN\cite{yao2018mvsnet, murez2020atlas, sun2021neuralrecon}.
  Moreover, geometry priors are leveraged to solve problems of noisy depth values\cite{zhang2018deep} and sparse depth input\cite{cheng2018depth}.
  However, there are still drawbacks to these methods.
  The use of CNN forces them to represent scenes discretely, creating limited precision and huge memory cost.
  Also, the domain gap can be a constraint to their generalization ability.

\paragraph{Implicit Neural Representation of Geometry}
Representing geometry with implicit neural networks has recently gained much attention for its high spatial resolution and low memory cost.
  In contrast to traditional 3D representations, it models scenes with a continuous function containing attributes such as volume density, signed distance, or occupancy in its 3D coordinate.
  With the help of geometry labels of spatial coordinates, implicit representation can be directly learned with global inputs\cite{park2019deepsdf,mescheder2019occupancy, chen2019learning}.
  However, it is hard to obtain ground-truth labels in real-world scenes. Therefore, researchers have put forward several methods to train the network without the strict requirements for labels. For example, SAL\cite{atzmon2020sal} and SA-ConvONet\cite{tang2021sa} leverages signed agnostic learning and Controlling Neural Level Sets\cite{atzmon2019controlling} takes another method to control the decision boundary directly.
  Furthermore, other methods\cite{chabra2020deep, jiang2020local} propose to decompose shape into local parts and use voxel-grids to store scene information.
  Optimization of the implicit function can also be reached by local supervision using color or depth images.
  NeRF\cite{mildenhall2020nerf} presents a volume rendering method to supervise implicit representation by 2D photometric loss.
  However, the use of the volume density has caused rough surfaces.
  To this end, methods introducing volume rendering into SDF\cite{yariv2020multiview, wang2021neus, ortiz2022isdf, yariv2021volume} or occupancy\cite{oechsle2021unisurf,zhu2021nice} representations have been proposed and achieves better surface reconstruction.
  Other methods fuse depth information into the frame work to further constraint optimization\cite{roessle2021dense, azinovic2021neural, deng2021depth, rematas2021urban}.

  However, as we have mentioned before, such methods either assume known spatial information or require a suitable environment.
  In this paper, we hope to relax restrictions on practical applications and explore an implicit reconstruction method more suitable for various scenarios and usages, leveraging the use of LiDAR.

\section{METHOD}

\begin{figure}[b]
  \centering
  \vspace{-4mm}
  \includegraphics[width=0.48\textwidth]{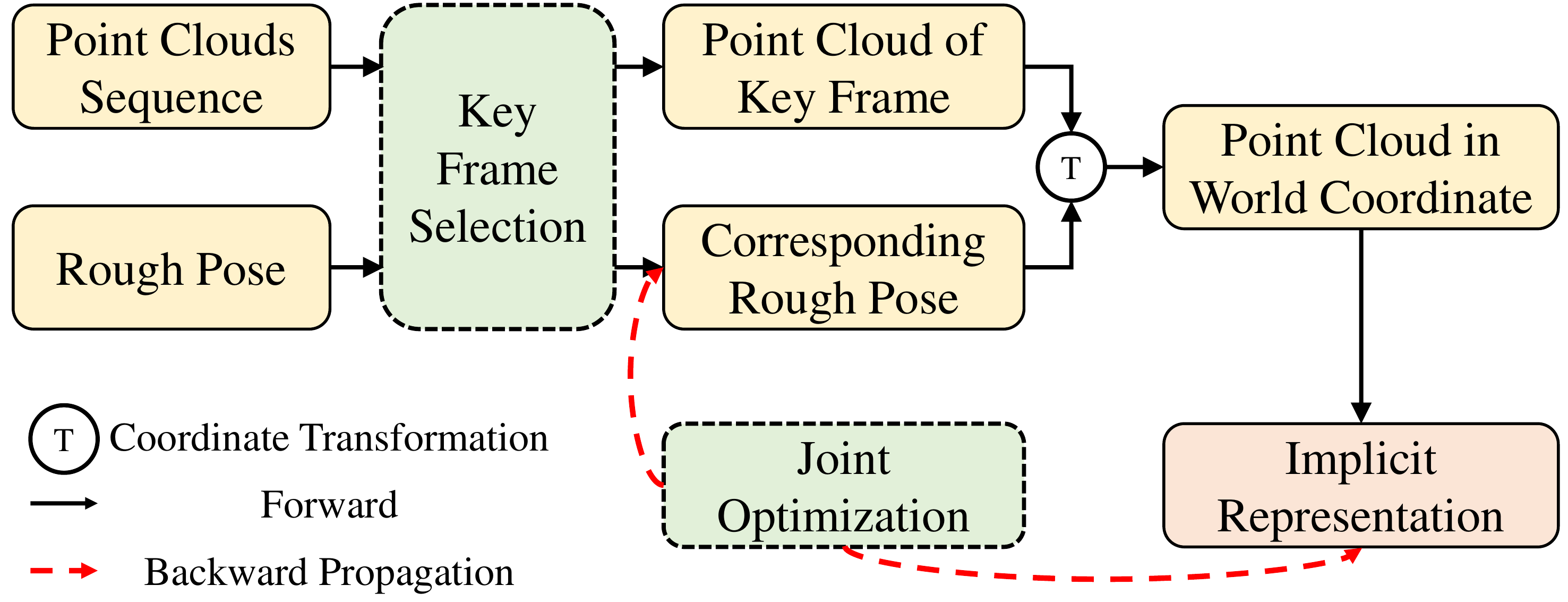}
  \vspace{-6mm}
  \caption{
    Pipeline of our method.
  }
  \label{flow_chart}
\end{figure}

\begin{figure*}[ht]
  \centering
  \includegraphics[width=0.83\textwidth]{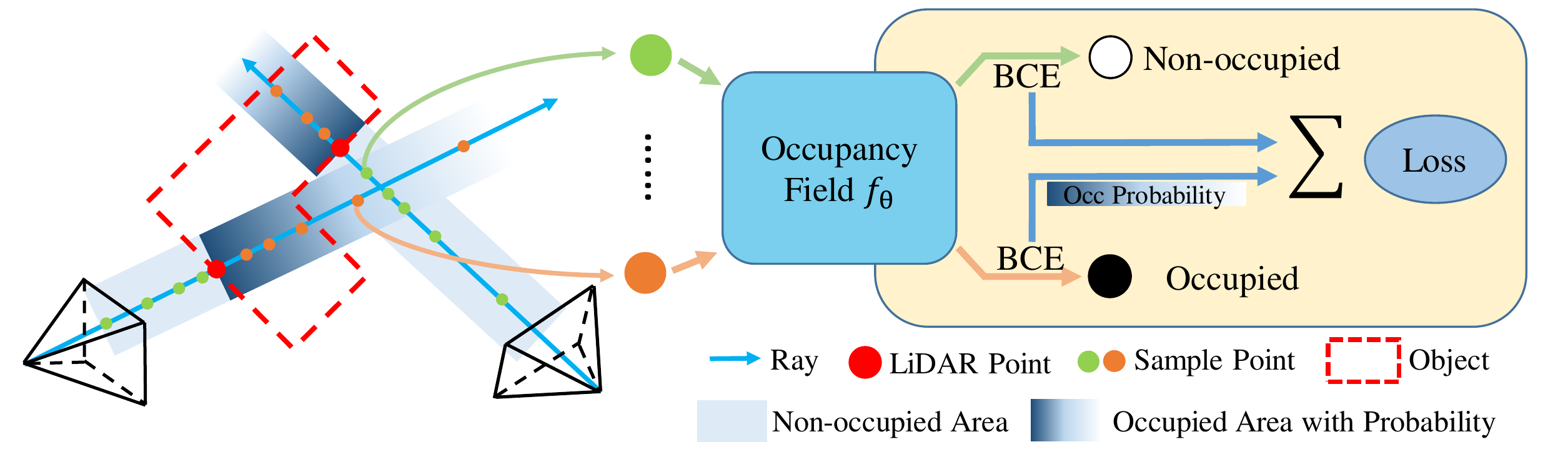}
  \vspace{-4mm}
  \caption{
    We use LiDAR depth measurement to separate a ray into two sides.
    We supervise the occupancy of sample points directly to occupied or non-occupied using a BCE loss.
    To make our direct loss occlusion aware, we add probability to occluded points using our thickness assumption.
    The final weighted BCE loss achieves unbiased and occlusion-awareness.
  }
  \label{thickness}
  \vspace{-6mm}
\end{figure*}

\subsection{System Overview}

In this work, we present a new implicit neural reconstruction method using sparse LiDAR point clouds.
  Our method takes point cloud frames as input and optimizes an occupancy field.
  We show the pipeline of our method in Fig. \ref{flow_chart}.
  Given input point cloud sequence and initial poses from odometry, we first select key frames according to the change of viewpoint for removing redundant frames and filter out poor depth measurements.
  We then sample points from the key frames and use them to train our implicit occupancy field.
  We use a multi-layer perceptron (MLP) with a multi-resolution hash encoder as our implicit model(Section \ref{implicit_representation}).
  Furthermore, We propose a direct supervision method to fully utilize depth information from sparse point cloud inputs (Section \ref{direct_supervision}).
  Finally, we jointly optimize our implicit network and sensor poses.(Section \ref{join_optimization}).
  In the end, the dense surface mesh is extracted from our optimized occupancy field using Marching Cubes\cite{lorensen1987marching}.

\begin{figure}[b]
  \centering
  \vspace{-5mm}
  \includegraphics[width=0.48\textwidth]{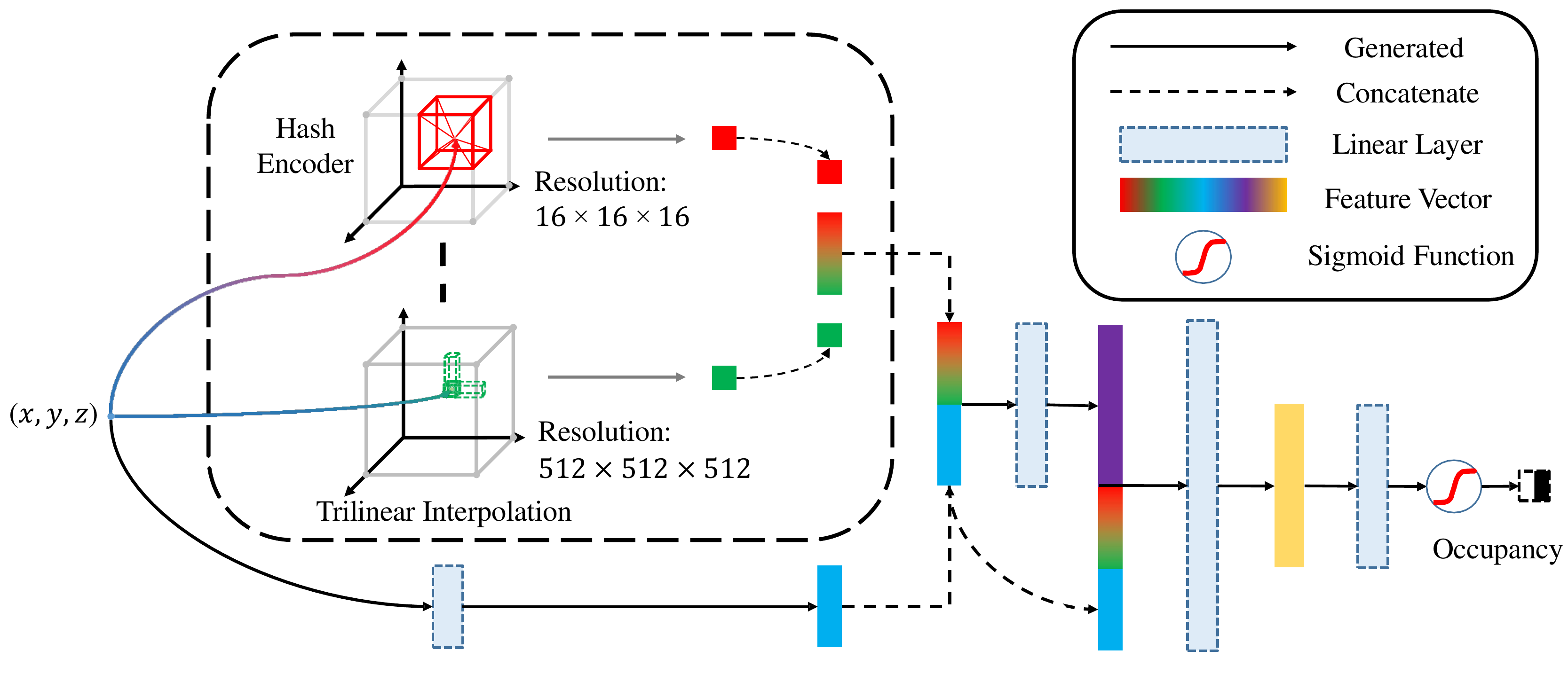}
  \vspace{-9mm}
  \caption{
        Implicit network architecture of our method.
  }
  \label{network}
\end{figure}

\subsection{Implicit Representation with Hash Encoder}
\label{implicit_representation}

Our network means to fit an implicit function ${f}_{\theta}(\mathbf{p}) : {R}^3 \rightarrow [0, 1]$ to the scene's occupancy field $\mathcal{O}(\mathbf{p}) : {R}^3 \rightarrow \{0, 1\}$, which maps a spatial coordinate to its occupancy property.
  Since we only have sparse point clouds as input, the supervision of the implicit function is relatively sparse too.
  Under such condition, a low-frequency-aware structure is needed to interpolate areas without supervision.
  On the other hand, high-frequency information also needs attention for a fine-grained reconstruction. 
  Therefore, we follow Instant-NGP and utilize a multi-resolution hash table to encode our coordinates to balance both low and high-frequency parts of the scene.
  
  The hash table is built upon multi-resolution voxel-grids of the scene.
  We use a spatial hash function to link learnable features with voxel coordinates.
  For a spatial point in the 3D space, its feature embedding can be obtained by trilinear interpolation.
  We then concatenate the feature vectors from all resolutions to form the frequency-aware multi-resolution latent vector. The use of the hash table can decrease memory cost to a great extent.
  However, hash collision may occur in high-resolution layers, which may cause artifacts in under-supervised areas.
  To avoid this, we extract an additional feature vector from the input coordinate and concatenate them as our final feature.
  
  Since the hash table has stored most scene features, we use a shallow MLP with low capacity to decode the features and get the occupancy decision boundary with a sigmoid function.
  Our complete network structure is shown in Fig. \ref{network}.
  
\subsection{Direct Supervision}\label{direct_supervision}

 In order to fully utilize the view information of each key frame, we treat points in the point cloud as rays.
  Specifically, given a point cloud frame containing $n$ points as $\{\mathbf{p}_i \left.\right| i=1, \cdots, n\}$, where each point is a vector of $(x_i, y_i, z_i)$ coordinate, the points on the $i$th ray can be represent as $\widetilde{\mathbf{p}}_i(z) = \mathbf{o}_i + z\mathbf{d}_i$, where $\mathbf{o}_i$ is the origin of the sensor and $\mathbf{d}_i$ is the direction of the ray.

  According to the mechanism of the LiDAR measurement, the laser can travel on the ray before hitting $\mathbf{p}_i$, which implies that the occupancy value along the ray before $\mathbf{p}_i$ should be $0$, and the value behind $\mathbf{p}_i$ should be $1$.
  In this way, a simple binary cross-entropy (BCE) loss can be directly applied to the occupancy function.
  \begin{align}\label{simple_BCE}
    \mathcal{L}_{d}(\widetilde{\mathbf{p}}_i(z)) = \left\{\begin{array}{lc}
      - \log (1-{f}_{\theta}(\widetilde{\mathbf{p}}_i(z))),
      &\text { for } z < z_i \\
      - \log({f}_{\theta}(\widetilde{\mathbf{p}}_i(z))), 
      &\text { for } z \geq z_i 
    \end{array}\right..
  \end{align}

 However, due to occlusion, we can not obtain object thickness and thus can not determine the range of the area be supervised as occupied.
  Simply treating the occluded space as occupied leads to ambiguity when the same region is observed by other views as non-occupied.
  
  To solve the occlusion problem and give supervision to points sampled behind objects, we make a simple object-thickness assumption.
  We propose a definition called generalized thickness.
  The generalized thickness $b$ refers to the distance of a single continuous occupied area that a ray $(\mathbf{o}, \mathbf{d})$ travels through:
  \begin{equation}
    b = min(z_s - z_i \left.\right| \mathcal{O}(\mathbf{o} + z_s \mathbf{d}) = 0.5),
  \end{equation}
 where $\mathcal{O} = 0.5$ means the level-set of the surface and $z_s$ means the depth of the very next surface the ray intersects with.
  Based on this definition, we treat the generalized thickness of objects in the environment as a random variable $B$.
We assume that it obeys a logarithmic normal distribution: $\ln(B) \sim \mathcal{N} (\mu, \sigma^2)$.
  Relying on this prior information of the scene's geometry, the probability of a point $\widetilde{\mathbf{p}}_i(z)$ sampled on the ray with $z\in(z_i, z_s)$ being occupied can be derived
  \begin{equation}
    P_{occ}(z) = P(B > z - z_a) = 1 - F_B(z - z_a),
  \end{equation}
 where $F_B$ means the cumulative distribution function of the variable $B$.
  Once the occupancy probability of the occluded sample point is obtained, supervision can be applied to the points behind the object.
  At the same time, a probability of $1$ is given to the non-occupied supervision.
  By weighting the simple BCE loss function with the probability of observation, an occlusion-aware direct loss on a point $\widetilde{\mathbf{p}}_i(z)$ can be modified as follows:
  \begin{align}
  \hspace{-2mm}
    \mathcal{L}_{d}(\widetilde{\mathbf{p}}_i(z)) = \left\{\begin{array}{lc}
      - \log (1-{f}_{\theta}(\widetilde{\mathbf{p}}_i(z))),
      &\text{for } z < z_i \\
      - P_{occ}(z)\log({f}_{\theta}(\widetilde{\mathbf{p}}_i(z))), 
      &\text{for } z \geq z_i 
    \end{array}\right.
  \end{align}
 
  With the occlusion-aware direct loss, the ambiguity can be solved, and a clean implicit neural representation can be optimized.

\begin{figure*}[t]
  \centering
\includegraphics[width=0.83\textwidth]{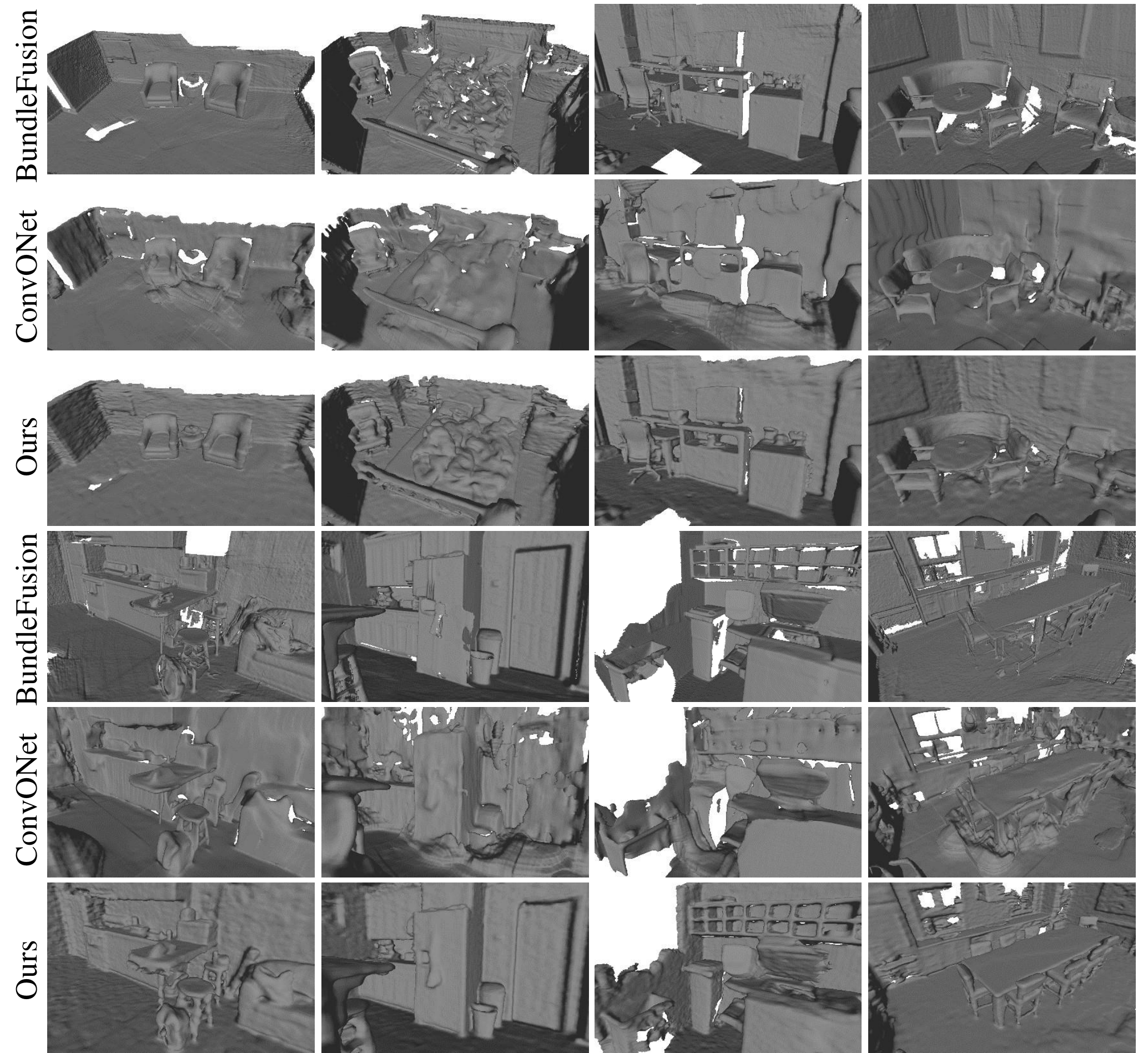}
  \vspace{-2mm}
  \caption{
    Qualitative results compared with BundleFusion and ConvONet on SacnNet dataset.
  }
  \label{scannet_result}
  \vspace{-2mm}
\end{figure*}

\begin{figure*}[t]
  \centering
  \includegraphics[width=0.84\textwidth]{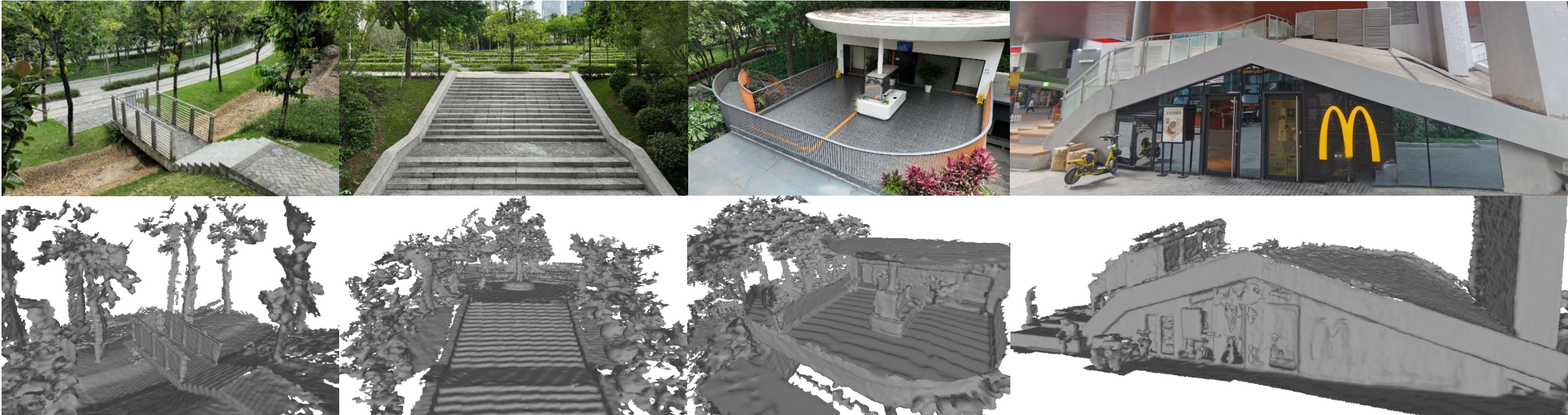}
  \vspace{-2mm}
  \caption{
    Qualitative results in large-scale outdoor environment.
    We show that our method is adaptable to various scenarios.
  }
  \label{self_collected_data}
  \vspace{-5mm}
\end{figure*}
\subsection{Joint Optimization}
\label{join_optimization}

 As we already have the surface position on each ray, to focus more on the near surface areas, we use a normal distribution to sample $m$ points on each ray: $\{\widetilde{\mathbf{p}}_i(z_j) \left.\right| j=1, \cdots, m\},  z_j \sim \mathcal{N} (z_i, \sigma_s^2)$, where $\sigma_s$ is a hyper parameter related to scene size.
  We additionally sample another $k$ points in a stratified way.
  During optimization, we first sample a batch of $N_b$ points from $N_f$ point clouds and form $N_f \times N_b$ rays.
  Then we sample $m+k$ points on each ray and use these total $ N_f \times N_b \times (m+k)$ points for supervision.

  Besides the loss function $\mathcal{L}_{d}(\widetilde{\mathbf{p}}_i(z))$ mentioned in Section \ref{direct_supervision}, we add another surface regularization loss $\mathcal{L}_{n}(\mathbf{p}_i)$ on surface points to encourages a smooth surface.
  The normal loss we use has the following format:
  \begin{equation}
    \mathcal{L}_{n}(\mathbf{p}_i) = \left| 1-
    \mathbf{n}(\mathbf{p}_i) \cdot \mathbf{n}(\mathbf{p}_i + \epsilon) \right|,
  \end{equation}
  where $\epsilon$ is a small neighbor range and $\mathbf{n}(\mathbf{p}_i)$ represents the normal of the implicit function at $\mathbf{p}_i$, denoted as
  \begin{equation}
    \mathbf{n}(\mathbf{p}_i)=\frac{\nabla_{\mathbf{p}_i} f_{\theta}(\mathbf{p}_i)}
    {\left\|\nabla_{\mathbf{p}_i} f_{\theta}(\mathbf{p}_i)\right\|_{2}}.
  \end{equation}
  Our final loss function is defined as
  \begin{equation}
    \mathcal{L} = \frac{1}{N_fN_b} \sum_{i=0}^{N_fN_b} \left( 
      \frac{1}{m+k} \sum_{j=0}^{m+k} \lambda_d \mathcal{L}_{d}(\widetilde{\mathbf{p}}_i(z_j)) + 
      \lambda_n \mathcal{L}_{n}(\mathbf{p}_i)
    \right).
  \end{equation}
  
  Since rays are transformed to world coordinate by poses of each frame, we can trace the gradient using backpropagation.
  Then we refine the initial poses while optimizing the implicit representation simultaneously by minimizing $\mathcal{L}$.
  This way, errors of the rough poses generated by the odometry system can be further reduced, creating a more consistent 3D geometry.

\begin{figure*}[!h]
  \centering
  \begin{overpic}[width=0.84\textwidth]{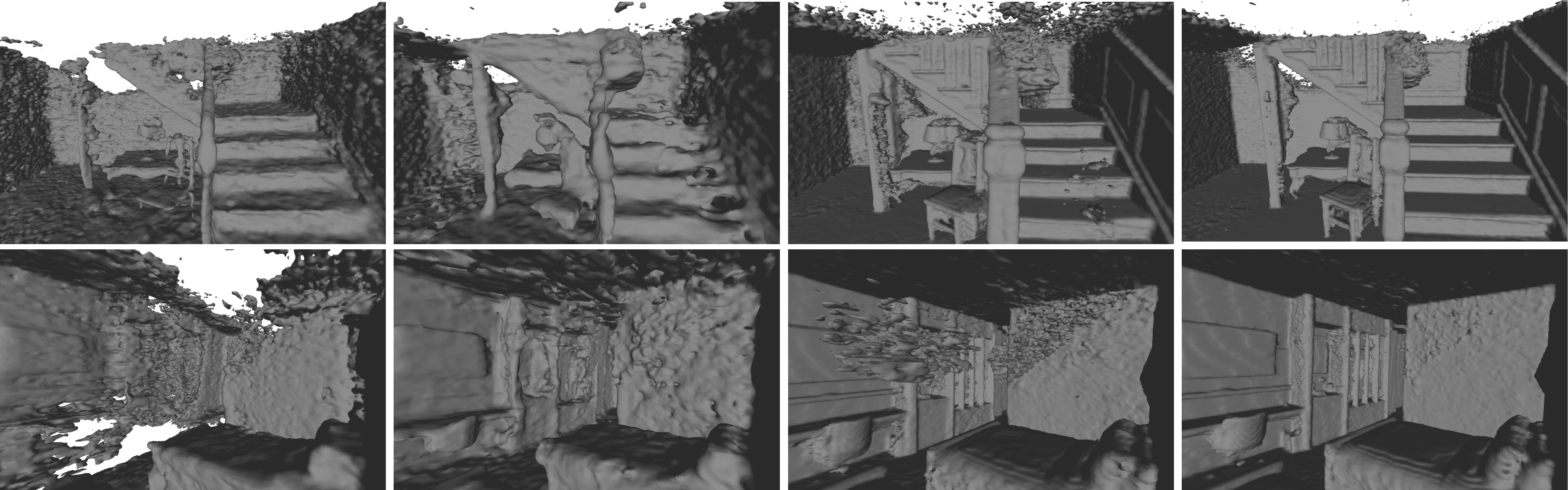}
    \put(4,-2.5){Depth render loss}
    \put(27,-2.5){w/o Pose refinement}
    \put(53,-2.5){w/o Thickness prior}
    \put(80,-2.5){Ours full model}
  \end{overpic}
  \vspace{2mm}
  \caption{
    Quantitative results of ablation studies.
  }
  \label{ablation_figure}
  \vspace{-1mm}
\end{figure*}
\section{EXPERIMENTS}

\subsection{Implementation Details}
  We implement our hash encoder following the setting in Instant-NGP.
  During optimization, we use ADAM\cite{kingma2014adam} optimizer with a learning rate of $1 \times 10^{-3}$ for both the implicit function and poses.
  The loss weight is set to $\lambda_d = 1.0$ and $\lambda_n = 0.4$.
  Every iteration, we sample points for supervision using $N_b = 100$, $m = 32$ and $k = 8$.
  The value of $\sigma_s$ is set to $0.3$ which is suitable for most scenarios.
  We run $300$ iterations per scene with around $5$ minutes on a single NVIDIA RTX2080Ti GPU.
  After $150$ iterations, we decrease the learning rate of pose refinement to $1 \times 10^{-4}$ to focus more on reconstruction.
  After optimization, we discretize our implicit function into voxel-grids of resolution $512$ and extract mesh using Marching Cubes.
  To cull the mesh in the unseen area, we filter out vertices $0.1$ meters away from our registered point cloud.

\subsection{Experimental Settings}\label{baseline}

\paragraph{Datasets}
  We perform quantitative evaluation on synthetic datasets, including 9-Synthetic-Scenes \footnote{The other 9 scenes tested in Neural RGB-D except ICL data.} \cite{azinovic2021neural} and Replica \cite{straub2019replica}.
  We also test our method on real-world datasets of ScanNet\cite{dai2017scannet} and our self-collected outdoor dataset.

\paragraph{Baselines}

  (1)\emph{COLMAP:} We run COLMAP\cite{schonberger2016structure} to generate poses and register a point cloud from depth images.
  We then use Screen Poisson surface reconstruction\cite{kazhdan2013screened} to generate mesh.
  (2)\emph{BundleFusion:} We feed dense color and depth maps directly to BundleFusion\cite{dai2017bundlefusion} for poses and geometric reconstruction.
  (3)\emph{ConvONet:} We use poses generated by BundleFusion and accumulate dense point clouds from the input depth map.
  We then run Convolutional Occupancy Network\cite{peng2020convolutional} on the global point cloud using a pre-trained network provided by the authors.
  (4)\emph{NeRF with depth loss:} We add additional depth loss to NeRF\cite{mildenhall2020nerf}, where rendered depth is compared with input depth map using L2 distance.
  The surface is extracted on the volume density field.
  (5)\emph{Neural RGB-D:} We use dense color and depth maps to perform Neural RGB-D \cite{azinovic2021neural} scene reconstruction.

\subsection{Results on Synthetic Datasets}
\label{synth_res}

\begin{table}[b]
\centering
\vspace{-4mm}
  \caption{
    Quantitative results on 9-Synthetic-Scenes dataset.\protect\\
    For fairness, metrics with * remove "the Kitchen" scene at evaluation for corruption data.
  }
\vspace{-2mm}
\begin{tabular}{cccccc}
\hline
\textbf{Method}       & \textbf{Sensor} & ${\textbf{C}-\ell_1}^*$   & $\textbf{F-score}^*$ & $\textbf{C}-\ell_1$   & $\textbf{F-score}$ \\
\hline
\textbf{COLMAP}       & RGB-D           & 0.036          & 0.835            & 0.060          & 0.743            \\
\textbf{BundleFusion} & RGB-D           & 0.046          & 0.809            & 0.067          & 0.788            \\
\textbf{ConvONet}     & RGB-D           & 0.050          & 0.680            & 0.073          & 0.658            \\
\textbf{NeRF-D}       & RGB-D           & 0.045          & 0.782            & 0.070          & 0.762            \\
\textbf{NeuralRGB-D}  & RGB-D           & \textbf{0.023} & \textbf{0.941}   & 0.048          & 0.917            \\
\textbf{Ours}         & \textbf{LiDAR}  & 0.030          & 0.921            & \textbf{0.030} & \textbf{0.920}  \\
\hline
\end{tabular}
\label{quantitative_eval}
\vspace{-2mm}
\end{table}

 We perform quantitative evaluation on 9-Synthetic-Scenes dataset.
  We evaluate the reconstructed mesh against ground truth using Chamfer $\ell_1$ distance and F-score\cite{knapitsch2017tanks} metrics with the threshold of 0.05 meter.
  We use the settings mentioned in Section \ref{baseline} following Neural RGB-D.
  To simulate a sparse LiDAR input, we randomly sample 15000 points per frame from the ground truth depth value and treat them as input data.
  We then use BundleFusion to generate initial poses.
  As shown in Table \ref{quantitative_eval}, our method receives comparable results with only sparse input (about $5\%$ of a depth map of $640 \times 480$ resolution).

  We also evaluate the efficiency of our method, and the results are shown in Table \ref{time_evaluation}.
  On average, our method requires fewer training frames and sparser rays per frame, resulting in less training time than Neural RGB-D while performing comparably.
  With our key frame selection scheme, hash encoder, and direct loss, we have saved training time to a great extent.

\begin{table}[h]
\centering
\vspace{-2mm}
  \caption{
    Time consumption results.
  }
\vspace{-2mm}

\begin{tabular}{ccc}
\hline
\textbf{Method}                   & \textbf{Neural RGB-D} & \textbf{Ours}  \\
\hline
\textbf{Average Frames Used}      & 1219                  & 118            \\
\textbf{Rays Per Frame}      & 307200                  & 15000            \\
\textbf{Average Time}   & 540 min               & \textbf{5 min}          \\
\hline
\textbf{Time Per Frame} & 26.6 s                & \textbf{2.5 s} \\
\hline

\end{tabular}

\label{time_evaluation}
\vspace{-2mm}
\end{table}

\begin{figure}[t]
  \centering
  \includegraphics[width=0.48\textwidth]{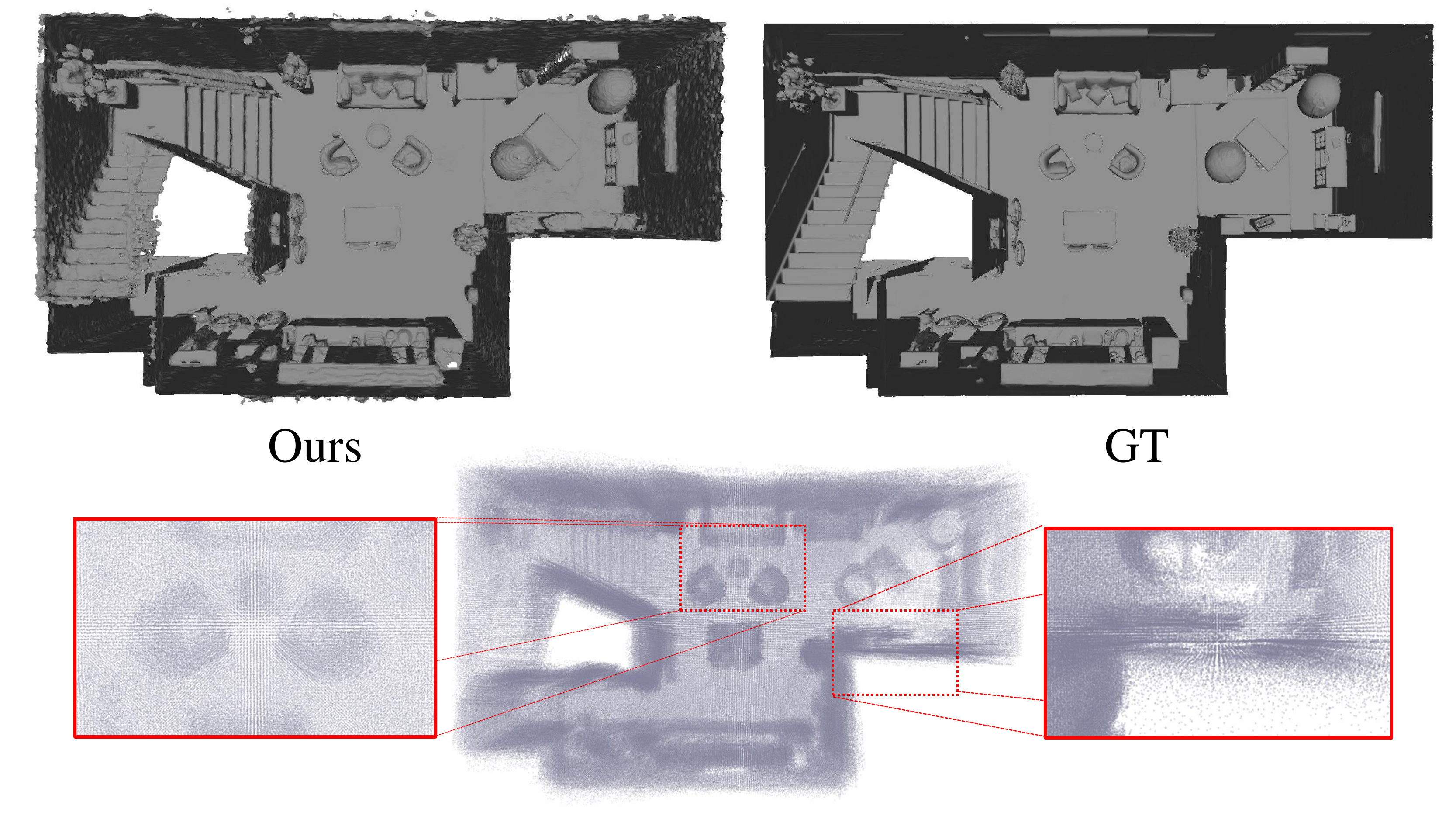}
  \vspace{-6mm}
  \caption{
    Qualitative results on Replica dataset.
    We evaluate the pose refinement ability of our method by giving noisy initial poses.
    The bottom image shows the accumulated global point cloud, which is blurry due to inaccurate initial poses.
  }
  \label{replica}
  \vspace{-6mm}
\end{figure}

  We test the pose refinement ability of our method on eight synthetic indoor scenes from the Replica dataset.
  Poses are initialized by adding noise to the ground truth.
  We add uniform distribution of $U(-0.05, 0.05)$ radius to rotation and $U(-0.1, 0.1)$ meter to translation.
  The generated mesh result can be seen in Fig. \ref{replica}.
  It shows that our direct supervision can optimize poses with noise and achieves consistent results.

\begin{table*}[t]
  \centering
  \vspace{-1mm}
    \caption{
    Quantitative results of ablation studies.
  }
  \vspace{-2mm}
  \begin{tabular}{ccccccccc}
  \hline
  \textbf{Method}  & \multicolumn{2}{c}{depth render loss} & \multicolumn{2}{c}{w/o pose refinement} & \multicolumn{2}{c}{w/o thickness prior} & \multicolumn{2}{c}{ours full model} \\ \hline
  \textbf{Metrics}  & \multicolumn{1}{c}{\textbf{C}-$\ell_1$}   & \textbf{F-score}  & \multicolumn{1}{l}{\textbf{C}-$\ell_1$}    & \textbf{F-score}   & \multicolumn{1}{c}{\textbf{C}-$\ell_1$}    & \textbf{F-score}   & \multicolumn{1}{c}{\textbf{C}-$\ell_1$}  & \textbf{F-score} \\ \hline
  9-Synth  & \multicolumn{1}{c}{0.098}      & 0.514        & \multicolumn{1}{c}{0.044}       & 0.553         & \multicolumn{1}{c}{0.036}       & 0.867         & \multicolumn{1}{c}{\textbf{0.030}}     & \textbf{0.920}       \\
  Replica  & \multicolumn{1}{c}{0.155}      & 0.341        & \multicolumn{1}{c}{0.049}       & 0.693         & \multicolumn{1}{c}{0.030}       & 0.879         & \multicolumn{1}{c}{\textbf{0.025}}     & \textbf{0.934}     \\ \hline
  \end{tabular}
  \label{ablation_table}
  \vspace{-4mm}
\end{table*}
\subsection{Results on Real-World Datasets}

  We employ ScanNet to operate experiments on real-world data.
  Similar to Section \ref{synth_res}, we randomly sample $15000$ points on the depth map as the pseudo LiDAR input.
  In Fig. \ref{scannet_result}, we compare our method to the original BundleFusion reconstructions and ConvONet.
  It can be seen that our model can generate a continuous surface and fill the holes that appeared in the BundleFusion models.
  We also solve the misalignment problems in ConvONet by our pose alignment module and generate a fine-grained surface with sparse input.

  Apart from the public datasets, we manage to test our framework using a commercial LiDAR in real world.
  We build up an outdoor LiDAR scan dataset using Livox AVIA, which generates $24000$ points per frame at $10$ fps.
  We run Fast-LIO\cite{xu2021fast} to obtain initial poses.
  As shown in Fig. \ref{self_collected_data}, our method works well in large-scale outdoor scenes with great movement.
  In this way, our framework further proves its practicability and broad application prospect.

\subsection{Ablation Studies}

  We analyze the effect of each module in our framework, including direct loss, pose refinement, and thickness prior, by replacing or removing them from our framework one at a time.
  The complete results of the ablation study can be found in Fig. \ref{ablation_figure} and Table \ref{ablation_table}

\paragraph{Direct Loss vs. Depth Render Loss}
  As we have mentioned, direct loss enables us to supervise on sparse point clouds in much less training time and is one of the critical contributions of our framework.
  Here we replace the direct loss with the prevalent L2 depth rendering loss and compare the results in two synthetic datasets.

\paragraph{Effect of Pose Refinement}
  Sensor poses obtained from the odometry can be drifting and noisy.
  Directly using these poses will cause inconsistency in geometry.
  We test our method by evaluation after removing the pose refinement module.

\paragraph{Effect of the Thickness Prior}
 We analyze the effect of the thickness prior by removing the occlusion-aware term from the loss function and using the simple BCE loss (Equation \ref{simple_BCE}) to guide the training.

  Table \ref{ablation_table} shows that all three modules play an essential role in building our reconstruction framework.
  The direct loss function ensures that we can supervise the reconstruction well with sparse input data, which relaxes the strict requirements on input data and helps reduce the training time.
  The pose refinement module refines the initial poses jointly and ensures the continuity and accuracy of the final result.
  The thickness prior solves the ambiguity of occupancy in the occluded space and removes the artifacts.
  We believe that the experiments once again prove the rationality and innovation of our framework.

\section{CONCLUSION}

  In this paper, we put forward an implicit 3D reconstruction method using only sparse point cloud frames from a LiDAR.
  We use implicit representation with the hash encoder and a newly proposed direct loss to deal with the sparsity of input data and suggest a thickness assumption to handle the occlusion problem.
  The experiments show that with around $5\%$ of the input data, we can achieve comparable reconstruction results with the methods using dense input.
  Moreover, our framework requires only a few minutes for training, significantly reducing time consumption.
  The limitation of our method is  that the hand-tuned thickness prior lacks generalization.
  In the future work, we will explore to solve this by adapting learnable priors.









{
\bibliographystyle{IEEEtran}
\bibliography{citation}
}



\end{document}